\documentclass[10pt,twocolumn,letterpaper]{article}

\usepackage[pagenumbers]{cvpr} 

\usepackage[dvipsnames]{xcolor}

\definecolor{cvprblue}{rgb}{0.21,0.49,0.74}
\usepackage[pagebackref,breaklinks,colorlinks,citecolor=cvprblue]{hyperref}

\def\paperID{14985} 
\def\confName{CVPR}
\def\confYear{2024}

\title{Few-Shot Classification \& Segmentation Using Large Language Models Agent}

\author{
Tian Meng $^1$$^\dagger$\hspace{2em} 
Yang Tao $^2$$^\dagger$\hspace{2em}
Wuliang Yin $^1$$^*$
\\
$^1$University of Manchester\hspace{2em}
$^2$Mettler Toledo Safeline 
\\
$^\dagger$Joint First Author\hspace{5em}
$^*$Project Lead
}

\begin{document}
\maketitle
\begin{abstract}
The task of few-shot image classification and segmentation (FS-CS) requires the classification and segmentation of target objects in a query image, given only a few examples of the target classes. We introduce a method that utilises large language models (LLM) as an agent to address the FS-CS problem in a training-free manner. By making the LLM the task planner and off-the-shelf vision models the tools, the proposed method is capable of classifying and segmenting target objects using only image-level labels. Specifically, chain-of-thought prompting and in-context learning guide the LLM to observe support images like human; vision models such as Segment Anything Model (SAM) and GPT-4Vision assist LLM understand spatial and semantic information at the same time. Ultimately, the LLM uses its summarizing and reasoning capabilities to classify and segment the query image. The proposed method's modular framework makes it easily extendable. Our approach achieves state-of-the-art performance on the Pascal-5i dataset.
\end{abstract}    
\section{Introduction}
\label{sec:intro}

Few shot learning \cite{Finn2017-yz,Wang2021-hb,Sung_2018_CVPR,Sun_2019_CVPR,Oreshkin2018-qd} is a machine learning paradigm where only a small number of examples are provided, and requires the model to generalize well to new, unseen data. Compared with the conventional heavy data paradigm, it relieves the burden of data labeling and annotation, and it also extend the application scenarios, where large amount of data cannot be collected easily. In computer vision, few shot classification \cite{Wertheimer_2021_CVPR,liu-2020} and few shot segmentation \cite{Kaixin2019-m,Rakelly2018ConditionalNF} has been actively studied. Although the setups of these two task are similar, only a few works \cite{kang79distilling,Kang_2022_CVPR} related them together. In this paper, we aim to solve the few shot image classification and segmentation (FS-CS), which requires to provide the presence of each support class in the query image as well as their pixel-level segmentation.

Despite the rapid advancements in few-shot learning approaches, traditional models heavily rely on meta-learning strategies \cite{hos-2022,Jake2017-m,wang2021meta,nichol2018first} or transfer learning paradigms \cite{zhu-2023, weiss2016survey, zhuang2020comprehensive} that typically inherit information from a base dataset with abundant labeled data. These models are then fine-tuned or adapted using a small number of examples from novel classes. However, such approaches often suffer from overfitting, especially in the context of few-shot segmentation, which demands a high level of generalization from limited annotations. The transformer-based models \cite{vaswani2017attention, ganesh2021compressing, gil-2020}, particularly vision transformers \cite{dosovitskiy2020image, liu2021swin, Khan2022-ij}, have recently shifted the paradigm by modeling image regions through self-attention mechanisms, providing a flexible architecture capable of capturing long-range dependencies. Although these models have the potential for few-shot tasks, they are still bottlenecked by the requirement of substantial computation and fine-tuning on novel datasets. 

\begin{figure}[t]
  \centering
   \includegraphics[width=1.0\linewidth]{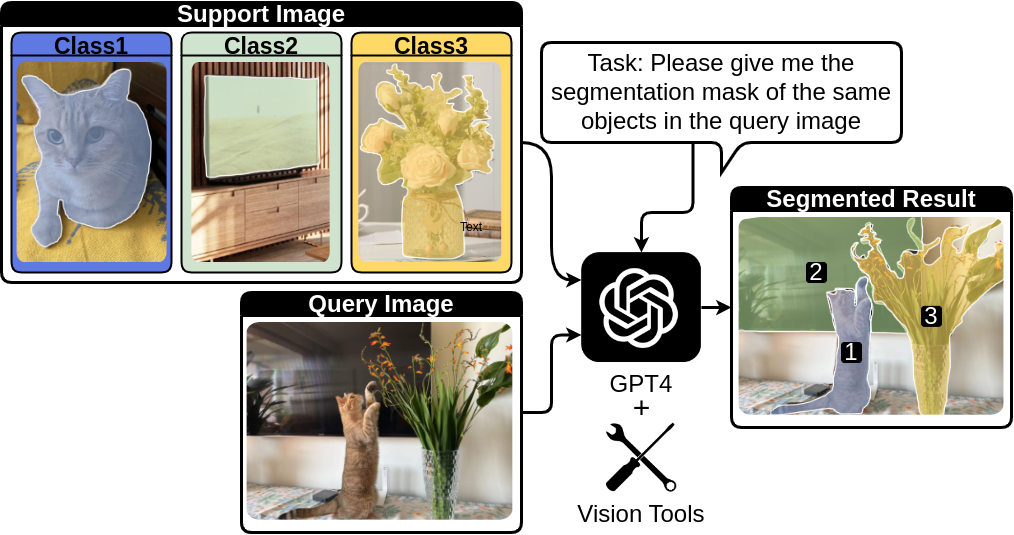}
   \caption{\textbf{Few-Shot Classification \& Segmentation Using Large Language Models Agent.} By providing vision tools to LLM like GPT-4, it can solve the task of Few-Shot Classification \& Segmentation with only image-level label in a training-free manner.}
   \label{fig:overview}
\end{figure}

Furthermore, the integration of language and vision to create multimodal models, such as CLIP \cite{radford2021learning} and DALL-E \cite{ramesh2021zero}, has opened new research avenues for leveraging descriptive language to guide visual understanding. While these advancements are encouraging, they have not fully utilized the potential of large language models (LLMs) in directly guiding few-shot image classification and segmentation (FS-CS) tasks. LLMs like GPT-4, with their in-context learning capabilities, can process complex instructions and understand abstract concepts, making them prime candidates to address FS-CS through a training-free approach. In this paper, we leverage the in-context learning and reasoning capabilities of a LLM to make sense of the visual world through textual descriptions and then guide pre-trained vision models, such as SAM \cite{kirillov2023segment} and GPT-4Vsion in performing FS-CS tasks. This strategy of LLM as an agent eliminates the need for extensive training or fine-tuning on novel datasets. 

To summarize, our contributions are the following:

\begin{itemize}
    \item We present a novel, modular framework that harnesses the power of LLMs for the FS-CS problem without additional training, enabling rapid adaptation to new tasks and domains.

    \item We exploit text prompting to guide the LLM in parsing few-shot tasks and generating action sequences for vision tools, and integrate text and visual prompting for VLM to perform visual tasks. This approach effectively transforms the LLM into an intelligent task planner that vision model tools for computer vision applications.

    \item We demonstrate state-of-the-art results on standard few-shot benchmark Pascal-5i \cite{shaban2017one}, showcasing the efficiency and effectiveness of our approach for both classification and segmentation tasks.
\end{itemize}

\section{Related work}
\label{sec:relatedwork}

\subsection{Few shot classification and segmentation}
Few-shot image classification and segmentation (FS-CS) \cite{kang79distilling,Kang_2022_CVPR} aims to generalize an algorithm to new classes not seen during training, given only a small sample of images. Few-shot classification \cite{Kang2021-ow,yin2021-m,gun2019-m} has been extensively studied, with approaches ranging from metric-based learning such as Siamese Networks \cite{Zhang_2019_CVPR} and Matching Networks \cite{Chang_2018_CVPR}, to model-based methods \cite{finn2017model} that meta-learn an internal model that can quickly adapt to new tasks. Few-shot segmentation \cite{dong2018few, wang2019panet}, on the other hand, has gained attention more recently. It expands the challenge by requiring pixel-level annotations of novel classes from a few instances. Approaches like prototypical networks \cite{NIPS2017_cb8da676}, which learn a metric space where segmentation can be performed as a form of nearest neighbor classification, and gradient-based meta-learning techniques \cite{NEURIPS2019_f4aa0dd9}, which aim at rapid adaptation of model parameters, have shown promising results. However, the generalization capability to new classes without further fine-tuning remains a substantial hurdle to overcome.

\subsection{Visual prompting}
Visual prompting is an innovative strategy that steers pretrained vision models' behavior using input modifications or augmentations \cite{jia2022visual, chen2023understanding, wu2022unleashing, zhang2023text}. It has roots in the success of textual prompts in LLMs and aims to achieve similar flexibility in vision tasks. Methods like CLIP \cite{radford2021learning} have pioneered the field by using text-image pairs to learn general visual features that can be queried using textual prompts. Visual prompting can be as simple as applying visual transformations that cue the model towards certain responses or as complex as synthesizing images. Our approach explores the efficacy of coupling textual and visual prompting for vision language models to enhance the problem-solving strategy delineated by an LLM. By conditionally adapting prompts based on the reasoning supplied by the LLM, we can direct the computational attention of vision models toward relevant aspects of the FS-CS tasks.

\section{Problem formulation}
\begin{figure*}
  \centering
  \includegraphics[width=1.0\linewidth]{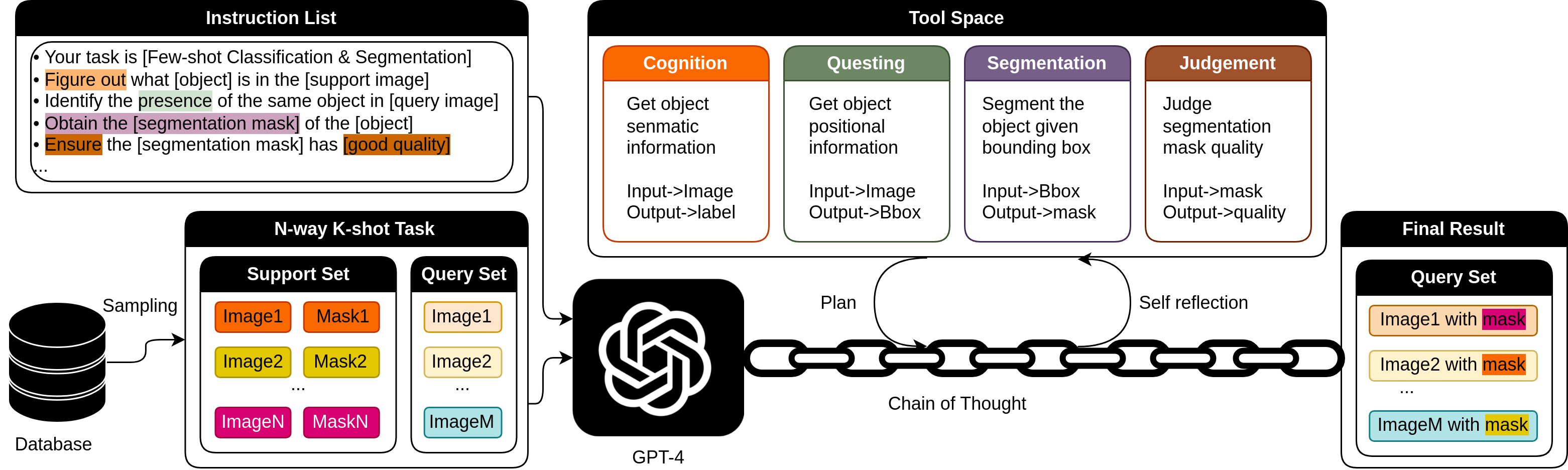}
  \hfill
  \caption{\textbf{Overview of LLM agent using visual tools to solve the task of few-shot classification and segmentation.} A TODO instruction list with a N-way K-shot task sampled from the database is provided to the LLM agent. Chain of thought is utilised to help LLM generating plan and sampling corresponded tools in the tool space. Self-reflection helps LLM improve the final segmentation result progressively.}
  \label{fig:short}
\end{figure*}

The core problem addressed in this paper is the Few-Shot Image Classification and Segmentation (FS-CS), which involves both identifying the category of target objects in a query image and delineating their precise pixel-level segmentation, given only a handful of labeled examples. Formally, we consider the few-shot setting with a support set \( S \) and a query set \( Q \). For a \(N\)-way \(K\)-shot classification and segmentation task, the support set \( S \) consists of \(N\) distinct object classes, each with \(K\) labeled examples: 
\begin{equation}
  S = \{(x_{n,k}, m_{n,k})\}_{n=1,k=1}^{N,K}
  \label{eq:supportDef}
\end{equation}
where \(x_{n,k}\) is the \(k\)-th image instance of the \(n\)-th class, and \(m_{n,k}\) is the binary mask for image \(x_{n,k}\), indicating the pixel-level presence of the \(n\)-th class object. \\
The query set \( Q \) is defined as:
\begin{equation}
  Q = \{(x', y', m')\} 
  \label{eq:queryDef}
\end{equation}
where \(x'\) is a new query image potentially containing instances of object classes from the support set \( S \), \(y'\) denotes the set of class labels present in \(x'\), and \(m'\) is the ground truth segmentation mask for the query image that needs to be predicted. The objective is to learn a model that uses the support set \( S \) to classify and segment the query image \(x'\) correctly.

\section{Method}
To tackle this problem, we propose a modular framework that divides the task into a sequence of sub-problems formalized through the use of Large Language Models (LLMs) as task planners, and then solve each sub-problem using corresponding tools.


\subsection{Task Planning}
Using the chain-of-thought approach, the LLM generates an explicit reasoning sequence that maps the support set \( S \) to a strategy that can be executed on any query image \(x'\) by vision models. The output is a structured action plan \( P \), containing a sequence of vision model tasks \(T\):
\begin{equation}
  P(T | S) = \text{LLM}_{\text{planner}}(S)
  \label{eq:taskPlanEq}
\end{equation}
This step effectively translates the problem into a set of instructions which can be understood and executed by vision models.
Based on the generated plan \( P \), vision models are prompted to perform the task \( T \), which involves classifying or segmenting the objects in the query image \(x'\). This step leverages pre-trained vision models that can understand spatial and semantic information, guided by the language-based plan:
\begin{equation}
  (\hat{y'}, \hat{m'}) = \text{execute}(T, x', \text{Vision Models})
  \label{eq:executeEq}
\end{equation}
where \(\hat{y'}\) is the predicted class label, and \(\hat{m'}\) is the predicted segmentation mask for the query image.
The effectiveness of the proposed framework is measured based on the accuracy of the classification and the quality of the segmentation for the query image \(x'\), with standard metrics such as Intersection over Union (IoU) for segmentation and accuracy for classification.
The novel contribution of this formulation is the decoupling of image understanding, task planning, and execution, allowing the LLM to use its reasoning capabilities without necessitating any end-to-end training or fine-tuning for novel FS-CS tasks. This approach aims at achieving rapid adaptation to new classes and tasks in a training-free manner, bridging the gap between natural language understanding and visual perception.

\subsection{Cognition}
\begin{figure}[t]
  \centering
   \includegraphics[width=1.0\linewidth]{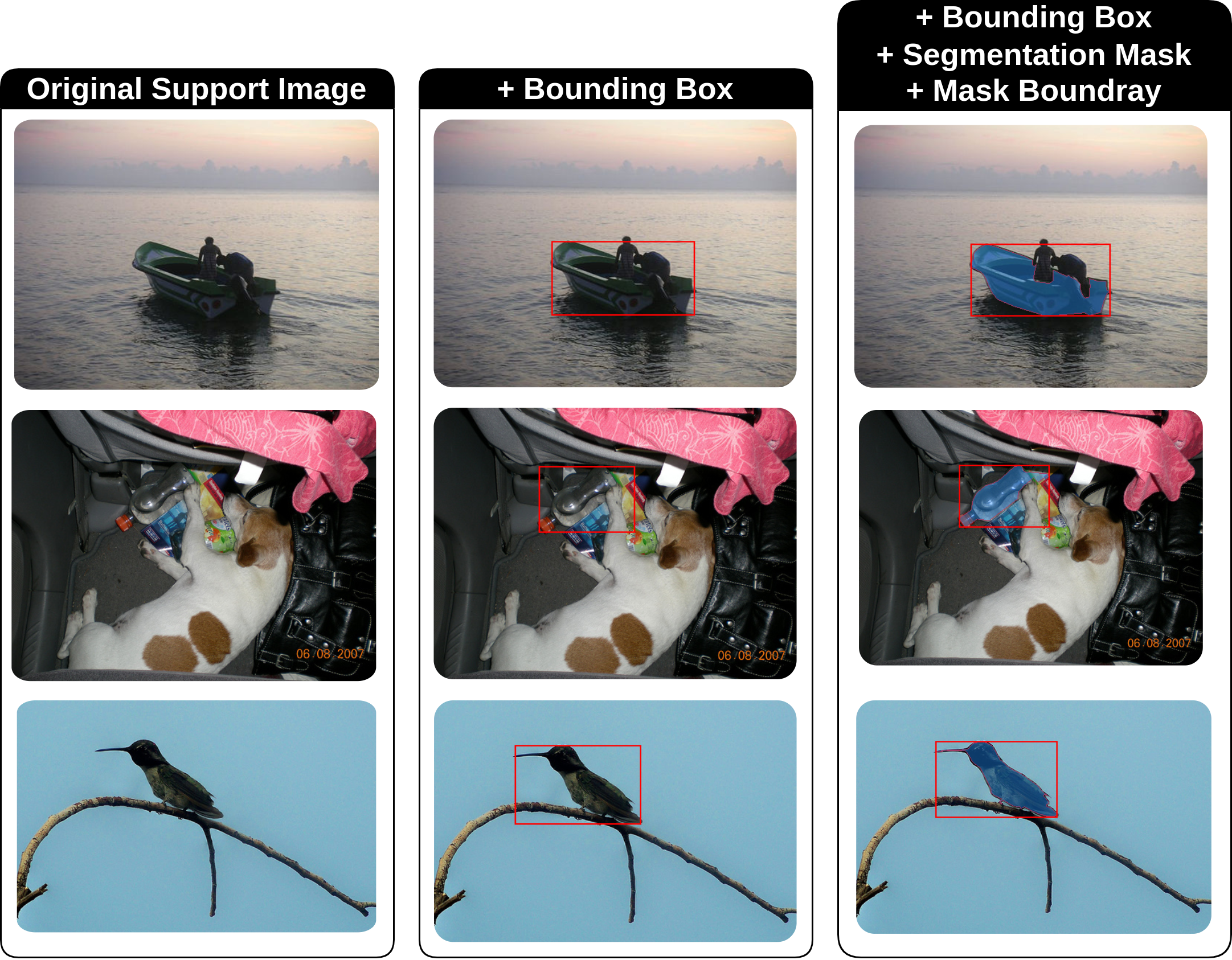}
   \caption{GPT-4Vision visual prompt for cognition}
   \label{fig:cognition}
\end{figure}

Cognition task processes the support set \( S \), recognizing the classes and the associated examples. It encompasses both identifying the target classes and extracting semantic insights from the support images. 
The cognitive task is formalized as follows:
\begin{equation}
  C = \text{execute}(\text{cognize}, S, \text{GPT-4Vision})
  \label{eq:cognitionEq}
\end{equation}
where \( C \) represents the cognitive output, containing a mapping between language-based descriptions and visual features that other tasks can use. \\
The GPT-4Vision's comprehension of natural language descriptions is harnessed to establish a cognitive connection between the textual class and their visual counterparts. These insights inform the construction of a task plan, which outlines how the vision models should interpret the query image \( x' \). A hybrid prompting, including both text and visual prompting, is utilised to allow GPT-4Vision to understand the target object in the support image and provide accurate language-based descriptions. The image-level labels and other possible metadata associated with the support set are provided to GPT-4Vision as shown in \cref{fig:cognition}. On the other hand, the coupled text prompting, including the descriptive information such as \textit{"The target object is in the RED bounding box and covered by LIGHT BLUE mask."}, is also used to guide GPT-4Vision to complete the task.

\subsection{Questing}
\begin{figure}[t]
  \centering
   \includegraphics[width=1.0\linewidth]{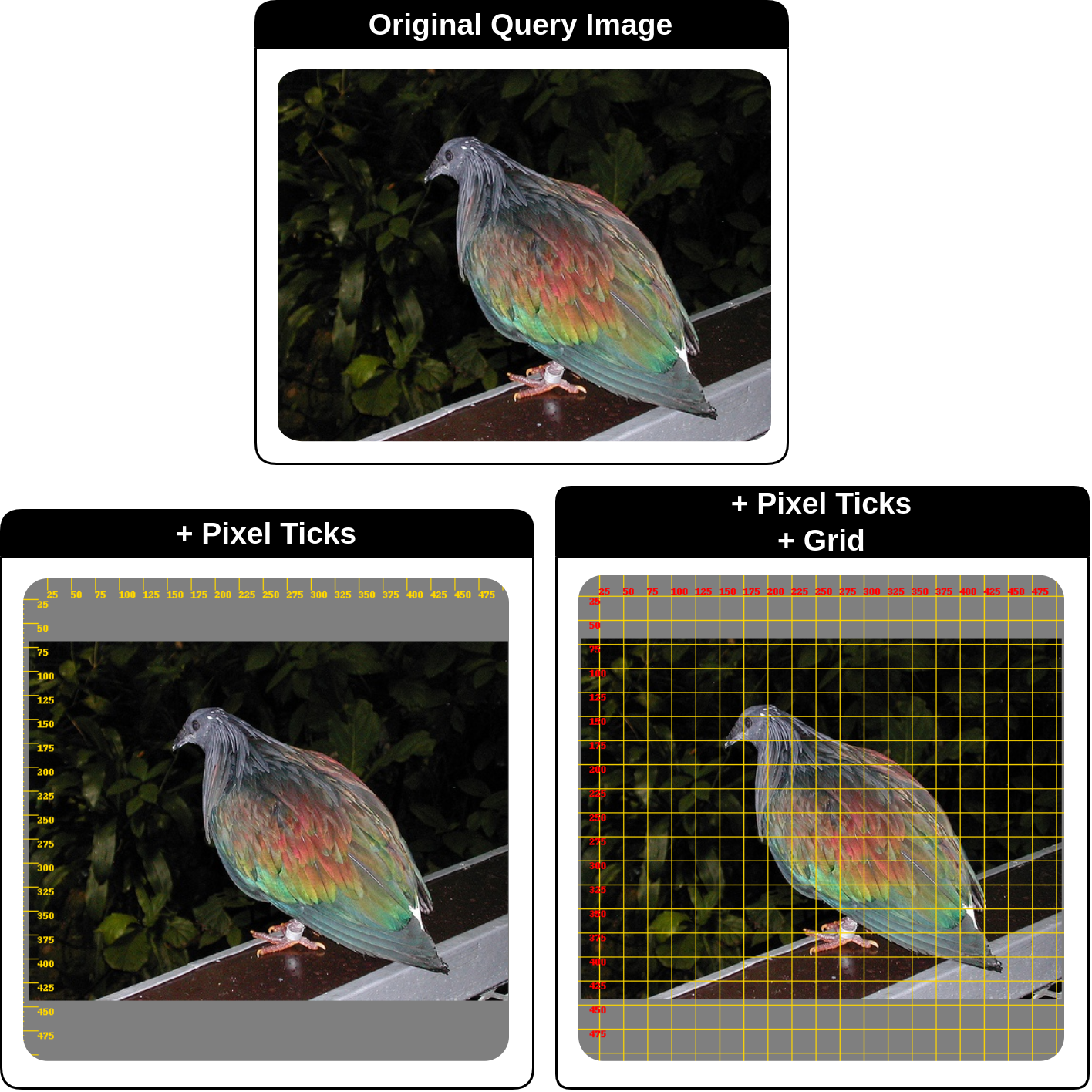}
   \caption{GPT-4Vision visual prompt for questing}
   \label{fig:questing}
\end{figure}

With cognitive outputs in hand, the LLM could employs the "questing" task where it utilises GPT-4Vision to perform specific tasks on the query image. Here, "questing" refers to the journey of seeking out and localizing target objects within the image. 
For each class in the support set, the LLM instructs GPT-4Vision to:
(i) Determine the presence of the same object within the query image \( x' \), giving rise to a binary decision for classification.
(ii) Provide the bounding box coordinates of the object assuming its presence, which serves as a precursor for the segmentation task.
This is described through the questing function:
\begin{equation}
  R = \text{execute}(\text{quest}, \{C, x'\}, \text{GPT-4Vision})
  \label{eq:questingEq}
\end{equation}
where \( R \) represents questing response, including class presence indicators and bounding box coordinates.
GPT-4Vision processes the cognitive insights \(C\) and the query image \(x'\), executing the dual objectives of identification and localization. GPT-4Vision's capability to interpret and adhere to the guidance of LLM-generated plans is crucial to the success of this step. Similarly, related visual and text prompting are provided to guide GPT-4Vision to output the bounding box accurately. We used a new style of visual prompting, as shown in \cref{fig:questing}, by plotting the coordinate ticks or grid directly on the image to aid object localization.

\subsection{Segmentation}
LLM provides the foundational input for the "Segment Anything model" (SAM), which specializes in creating pixel-level representations of objects within the given bounding boxes. SAM leverages the bounding box coordinates to mask the pertinent region of the query image, distinguishing between the target object and its background.
The segmentation process is specced out thus:
\begin{equation}
  \hat{m'} = \text{execute}(\text{segment}, \{R, x'\}, \text{SAM})
  \label{eq:segmentationEq}
\end{equation}
where \( \hat{m'} \) symbolizes the output segmentation mask for each class depicted in the query image \( x' \).
SAM's performance is instrumental, as it pushes the frontiers of few-shot segmentation by establishing class-specific segmentation without needing class-wise annotated examples, which is different from traditional methods relying on extensive fine-tuning with detailed pixel annotations.

\subsection{Quality Judgement and Self-Reflection}
The final stage in our framework involves a self-reflection task, designed to critically evaluate and improve the segmentation quality. The LLM re-engages post-segmentation to perform a binary assessment of \( \hat{m'} \).
The GPT-4Vision reviews the segmentation quality of \( \hat{m'} \) with codified qualitative metrics, such as shape conformity, coverage, and class confidence, to form a judgement on segmentation quality:
\begin{equation}
  J = \text{execute}(\text{judge}, \{C, \hat{m'}\}, \text{GPT-4Vision})
  \label{eq:judgementEq}
\end{equation}
where \( J \) denotes the judgement output, which is a binary approval and a set of refinement suggestions.
Should the quality not meet predefined standards, the self-reflection loop triggers iterative adjustments via the LLM. The LLM refines the plan \( P \), leading to a recalibration of both the questing \( R \) and segmentation \( \hat{m'} \). This iterative mechanism ensures continuous improvement and learning, essentially embodying a vigilant quality control overseer within the FS-CS framework. Through this self-aware process, the proposed method not only sustains high accuracy but also fosters an internal validation mechanism, promoting reliability and trustworthiness in autonomous FS-CS systems.\\
In addition, in context learning is utilised to guide GPT-4Vision gives objective assessment and constructive refinement suggestions. By providing examples of good and poor segmentation outcomes along with their critiques,
which enables GPT-4Vision to adaptively learn the judgement standard. And LLM could refine the segmentation mask over successive iterations.
\section{Experiment}
In order to validate the efficacy of our proposed framework, we conducted extensive experiments on the widely used few-shot learning benchmark Pascal-5i. We compared our model's performance against several state-of-the-art few-shot classification and segmentation approaches. This section details the experimental setup, datasets, evaluation metrics, baselines for comparison, and the results obtained.

\subsection{Datasets}
Pascal-5i derived from PASCAL VOC Challenge contains 20 object classes and is split into 4 different sets, with each set treated as one cross-validation fold. We follow the standard protocol for few-shot segmentation where each class is evaluated under the one-shot scenarios.
\begin{figure}[t]
  \centering
   \includegraphics[width=1.0\linewidth]{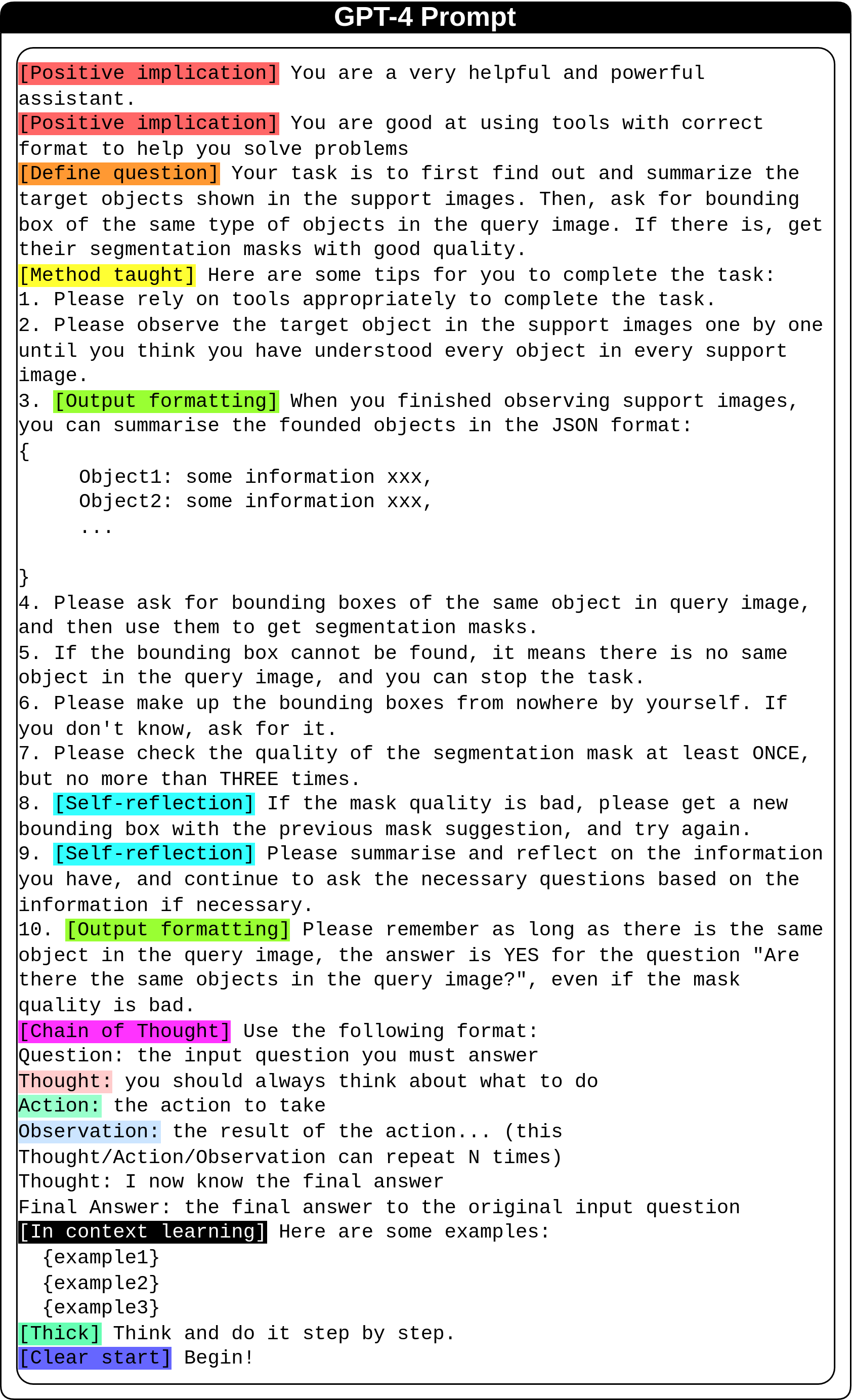}
   \caption{GPT-4 prompt for the task of Few-Shot Classification \& Segmentation}
   \label{fig:gpt4_prompt}
\end{figure}

\begin{figure}[t]
  \centering
   \includegraphics[width=1.0\linewidth]{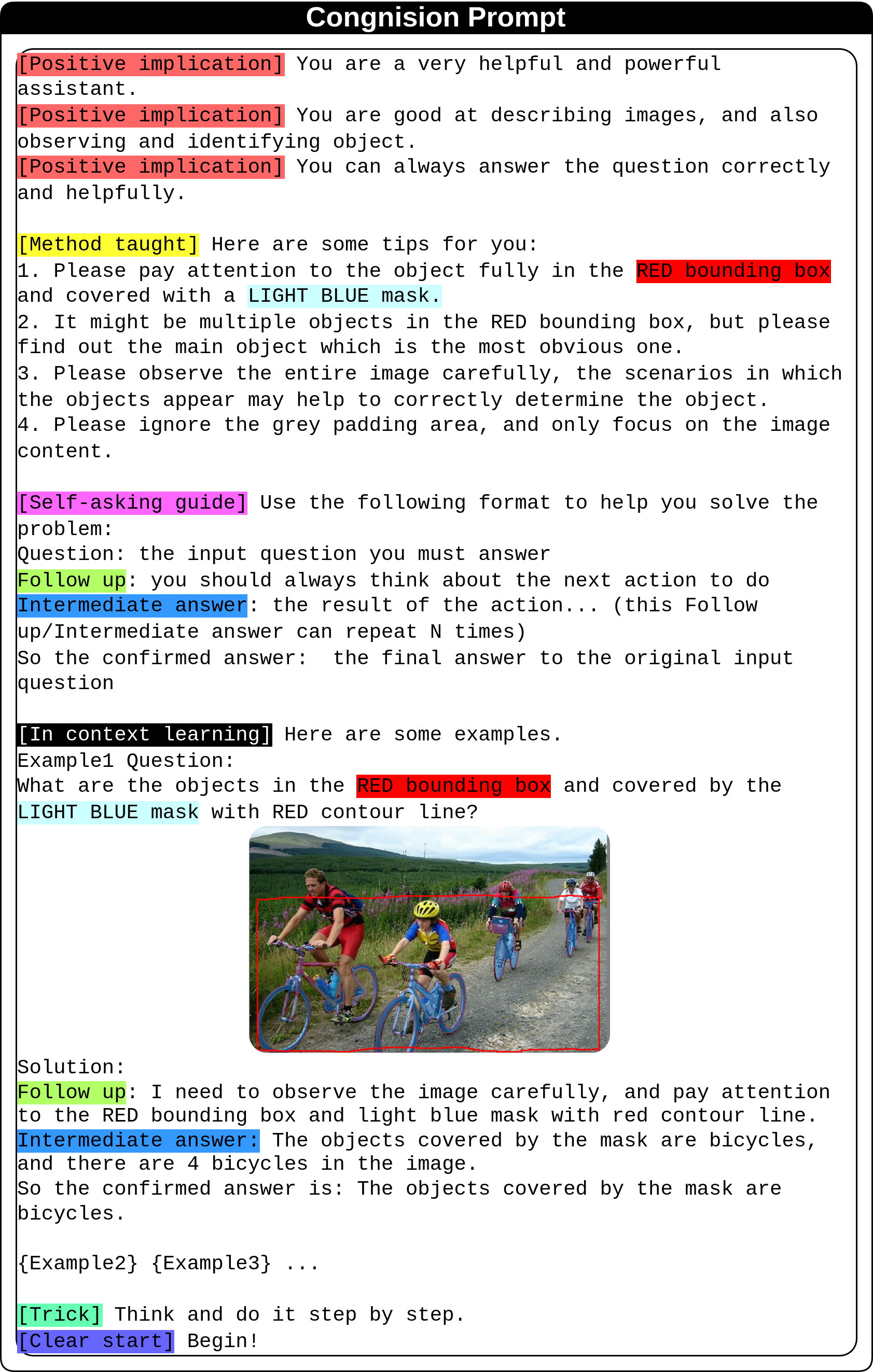}
   \caption{GPT-4Vision prompt for cognition object in support image}
   \label{fig:cognision_prompt}
\end{figure}

\begin{figure}[t]
  \centering
   \includegraphics[width=1.0\linewidth]{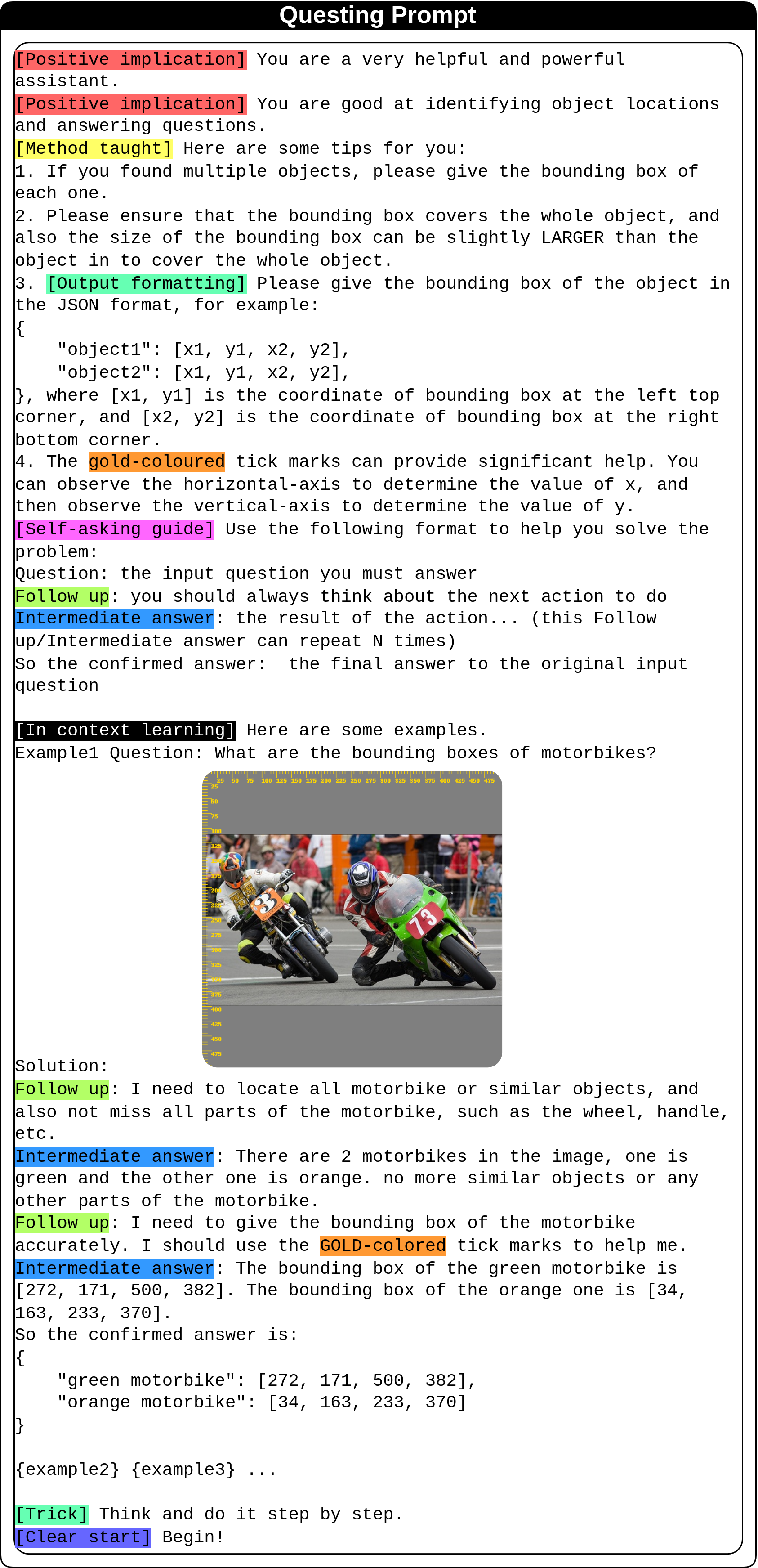}
   \caption{GPT-4Vision prompt for identifying the query image object and localisation}
   \label{fig:questing_prompt}
\end{figure}

\begin{figure}[t]
  \centering
   \includegraphics[width=1.0\linewidth]{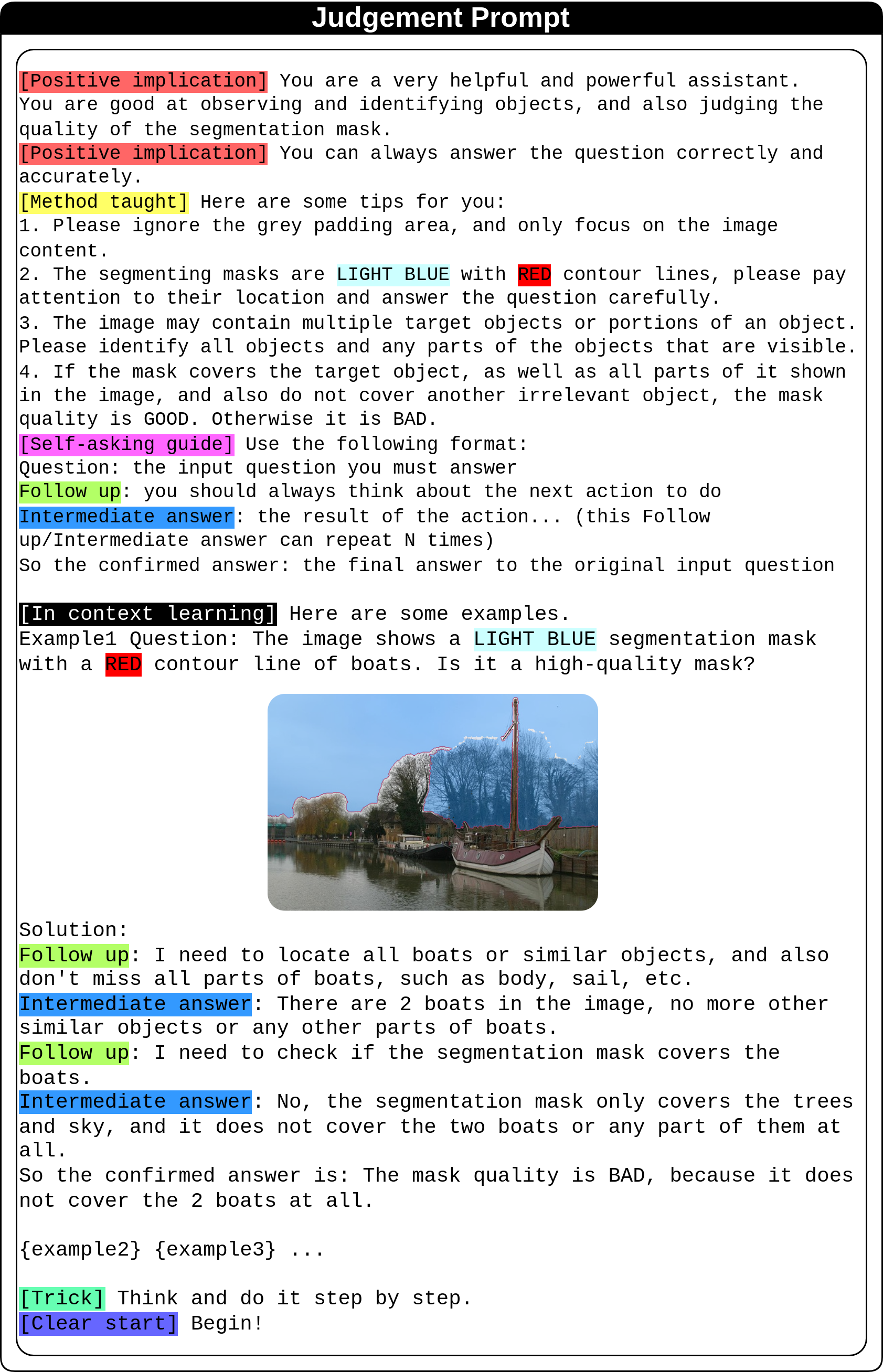}
   \caption{GPT-4Vision prompt for judging the segmentation mask quality}
   \label{fig:judgement_prompt}
\end{figure}

\begin{figure}[t]
  \centering
   \includegraphics[width=1.0\linewidth]{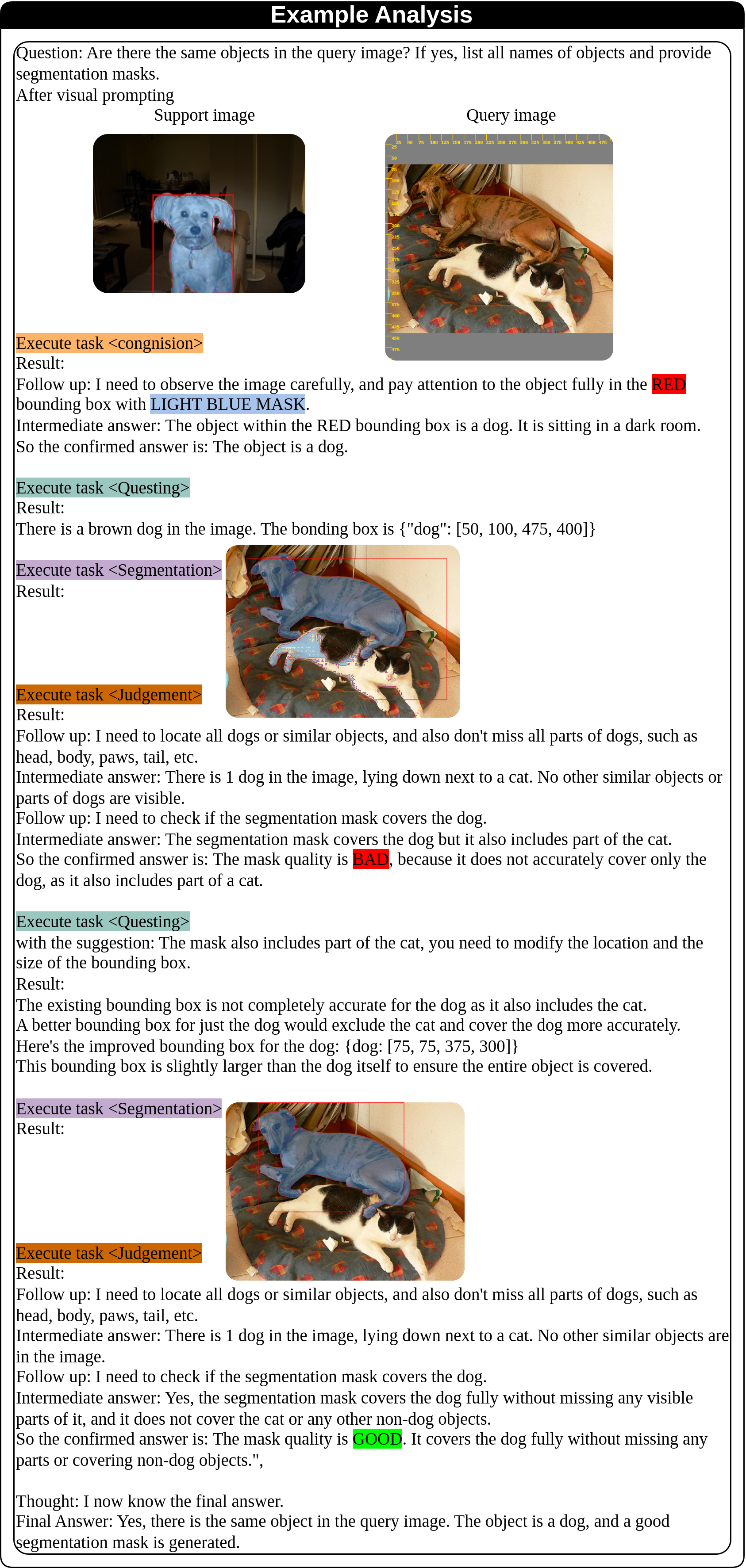}
   \caption{A example showing the workflow of solving the task of classicication \& segmentation}
   \label{fig:example_detail}
\end{figure}

\subsection{Implementation Details}
Our Large Language Model (LLM) used for task planning and reasoning is based on GPT-4, while the vision models involved are pre-trained versions of Segment Anything Model Huge (SAM-H) and GPT-4Vision. We apply a chain-of-thought prompting method to guide the LLM through the reasoning process while employing a hybrid form of visual and textual prompting for GPT-4Vision. Detailed prompt designs are described as the following.\\
\textbf{LLM task planning.} The purpose of the LLM task planning prompt, as shown in \cref{fig:gpt4_prompt}, is to articulate and organize the approach to the few-shot classification and segmentation task. By using a chain-of-thought style of prompting, we encourage GPT-4 to (i) Generate a step-by-step reasoning sequence reflecting the LLM's understanding of the task requirements. (ii) Formulate an action plan that indicates how vision models should be used to classify and segment objects in the query image, given the information about support classes. (iii) Reflect the obtained information and continue to improve segmentation mask quality. \\
\textbf{Cognition.} In the cognition prompt, as shown in \cref{fig:cognision_prompt}, the main aim is to leverage GPT-4Vision to interpret the visual data provided in the support set and establish cognitive links between the textual labels and their visual representations. The prompt guides GPT-4Vision to: (i) Recognize the target classes within the support images.
(ii) Describe the target object accurately leveraging any metadata provided. The cognitive prompt could be structured as part narrative and part directive, telling a story of the image and guiding GPT-4Vision to focus on certain attributes relevant to the task, such as distinguishing features of classes or peculiarities in the images that might aid classification and segmentation. \\
\textbf{Questing.} The questing prompt, as shown in \cref{fig:questing_prompt}, serves to direct GPT-4Vision to identify and localize objects in a new query image as specified by the LLM-generated plan. This prompt: (i) Instructs GPT-4Vision to search for instances of the support set classes within the query image. (ii) Guides GPT-4Vision to provide a binary classification decision on whether each class is present or not in the query image. (iii) Directs GPT-4Vision to discern and output bounding box coordinates for objects, serving as input for the segmentation task if a class is present. The questing prompt could utilize cues on how to perform visual search within the image, alongside the cognitive insights generated earlier, to improve GPT-4Vision's accuracy in object localization and identification. \\
\textbf{Judgement.} Following the output generation by the vision model, the judgement prompt, as shown in \cref{fig:judgement_prompt}, involves self-assessment by GPT-4Vision on the quality of the generated segmentation masks. The judgement prompt encourages GPT-4Vision to: (i) Critically review the output masks with respect to the input query image and cognitive output. (ii) Assess the segmentation quality using pre-defined criteria such as conformity to the object shape, coverage of the object area, and confidence of the classified segments. (iii) Offer binary approval of the segmentation quality and provide refinement suggestions if the quality does not meet the criteria. This prompt would be used to trigger a refinement loop, if necessary, based on the self-review, thus embodying a sort of internal quality control to ensure the reliability of the FS-CS system.
\begin{table*}
  \centering
  \begin{tabular}{@{}lclclclclclclclclclc@{}}
\toprule
 &\multicolumn{10}{c}{1-way 1-shot}\\
 & \multicolumn{5}{c}{classification 0/1 exact ratio (\%)}&  \multicolumn{5}{c}{segmentation mIoU (\%)}\\
 Method& $5^0$& $5^1$& $5^2$& $5^3$& avg.& $5^0$& $5^1$& $5^2$& $5^3$&avg.\\
\midrule
    HSNet& 84.5  & \textbf{84.8} & 60.8 & 85.3& 78.9& 20.0&  23.5&  16.2 & 16.6 &19.1\\
    ASNet& 80.2 & 84.0 & 66.2 & 82.7 & 78.3& 11.7&  21.1 & 13.4 & 16.2 &15.6\\
    DINO& -& -& -& -& -& 20.0&  23.4 & 16.2 & 16.6 &19.1\\
    CST& 84.0&  82.2 & 70.8 & 82.6 & 79.9& 35.8&  38.9 & 28.9 & 29.2 &33.2\\
    \textbf{Ours}& \textbf{93.5}& 80.3& \textbf{84.4}& \textbf{87.3}& \textbf{86.4}& \textbf{37.3}& \textbf{45.5}& \textbf{34.2}& \textbf{35.6}&\textbf{38.2}\\
 \bottomrule
  \end{tabular}
  \caption{Comparing model performance on FS-CS of th 1-way 1-shot setting.}
  \label{tab:accuracy}
\end{table*}

\subsection{Results}
For classification, we report the average classification accuracy across all classes and folds. For segmentation, Intersection over Union (IoU) is employed as the key metric. We calculate the mean IoU across all classes and query images to obtain the overall performance evaluation. The result is shown in \cref{tab:accuracy}. Our model achieved remarkable results, surpassing all baselines by a significant margin on Pascal-5i dataset. In one-shot classification, our model achieved an average accuracy of \(\text{86.4\%}\), outperforming the best baseline by \(\text{6.5\%}\). In the segmentation task, our model demonstrated superior performance with an average IoU of \(\text{38.2\%}\) in the one-shot scenario, surpassing the strongest baseline by \(\text{5.0\%}\). \\
\textbf{Detailed Example Analysis.} In this subsection, we zoom in on a specific one-shot instance from the Pascal-5i dataset to elucidate how our model functions in a practical setting. We selected an example, as shown in \cref{fig:example_detail}, that is challenging due to the presence of multiple object instances within the scene. The chosen query image depicts an scene with a cat and a dog. The support image provided was that of a dog. This example tests the model's capacity to recognize and segment the dog instances while discriminating against other objects. Here is how our model processed this example:
\begin{enumerate}
    \item Cognitive Observation. Upon reviewing the support image, GPT-4Vision provided a detailed description, recognizing the target object (a dog) in the bounding box and covered by the light blue mask. This cognitive output includes a description of the size, shape, spatial relationships between different parts.

    \item Questing and Localization. Equipped with the cognitive insights, GPT-4Vision commenced the questing task on the query image. The model successfully identified the dog from the cat and background. It determined the presence of the dog class, issued positive classification decisions, and provided bounding box coordinates at the first time.

    \item Segmentation. The SAM acted upon the bounding box coordinates provided by GPT-4Vision and yielded segmentation masks that delineated the dog from the rest of the image content. However, due to the bounding box size is too large, a part of cat is also segmented out. 

    \item Quality Judgement. Following segmentation, GPT-4Vision engaged in self-reflection and quality judgement. On evaluating the segmentation masks against the established qualitative metrics, it noted some regions of imprecision. Utilizing the judgement prompt, GPT-4Vision offered a BAD assessment result and constructive feedback for mask refinement.

    \item Refinement. GPT-4 continue to invoke questing task and provide the feedback to GPT-4Vision. GPT-4Vision refined the bounding box based on the feedback. Finally, the new mask passed the by the quality judgement and the task stopped.
\end{enumerate}
Through this detailed instance analysis, we highlighted the model’s integrated intelligence - its capability to undertake complex visual reasoning, classification, and segmentation without the need for any additional fine-tuning. Additionally, the model displayed a robust understanding of object features and contexts, facilitated effective communication between language and vision components, and upheld high-quality standards through self-assessment and refinement loops. Overall, this case serves as a testament to the dynamism and adaptability of our framework, confirming its potential to deliver state-of-the-art performance on few-shot classification and segmentation tasks.

\section{Conclusion}
In conclusion, our framework demonstrates a novel and effective approach to few-shot classification and segmentation by bridging the high-level reasoning capabilities of Large Language Models with the precise image interpretation of vision models. It achieves state-of-the-art performance without the reliance on extensive dataset-specific training, highlighting the power of chain-of-thought prompting, in-context learning, and self-reflection for rapid task adaptation and quality execution. It shows a potential that language models act as meta-learners, capable of directing and refining visual processing tasks.

{
    \small
    \bibliographystyle{ieeenat_fullname}
    \bibliography{main}
}


\end{document}